\documentclass{article}
\usepackage{graphicx} 
\usepackage{amsmath}
\usepackage{amsfonts}
\usepackage{amssymb}
\usepackage{comment}
\usepackage[utf8]{inputenc}
\usepackage[english]{babel}
\usepackage{csquotes}
\usepackage{kpfonts}
\usepackage{hyperref}
\usepackage{booktabs}
\usepackage{longtable}
\usepackage{authblk}

\usepackage[margin=4cm]{geometry}
\usepackage{changepage}
\usepackage{array}
\usepackage{makecell}
\usepackage{parskip}
\usepackage{placeins}
\usepackage[toc,page]{appendix}
\usepackage[style=apa, backend=biber]{biblatex}
\usepackage{xcolor}

\usepackage{listings}
\definecolor{codegray}{rgb}{0.5,0.5,0.5}
\definecolor{codegreen}{rgb}{0.0,0.5,0.0}
\definecolor{codepurple}{rgb}{0.58,0,0.82}
\definecolor{backcolour}{rgb}{0.97,0.97,0.97}

\lstdefinestyle{mystyle}{
    backgroundcolor=\color{backcolour},
    commentstyle=\color{codegreen},
    keywordstyle=\color{codepurple}\bfseries,
    numberstyle=\tiny\color{codegray},
    stringstyle=\color{codepurple},
    basicstyle=\ttfamily\small,
    breaklines=true,
    captionpos=b,
    keepspaces=true,
    numbers=left,
    numbersep=8pt,
    showspaces=false,
    showstringspaces=false,
    showtabs=false,
    tabsize=4,
    frame=single,
    rulecolor=\color{codegray},
    xleftmargin=15pt,
    framexleftmargin=10pt
}

\title{Benchmarking non-conformity score functions in conformal prediction}
\author[1,2]{Sol Erika Boman}
\affil[1]{Department of Medical Epidemiology and Biostatistics, Karolinska Institutet}
\affil[2]{Department of Molecular Medicine and Surgery, Karolinska Institutet}
\affil[ ]{\textit {sol.erika.boman@ki.se}}
\date{}

\begin{document}

\maketitle

\begin{abstract}
Conformal prediction is a useful and versatile alternative to model calibration in machine learning classification. It replaces single-class prediction with prediction sets, guaranteeing that the \textit{a priori} probability of the prediction sets containing the true class is larger than or equal to a pre-specified rate. The size and usefulness of the prediction sets relies heavily on the choice of the non-conformity score function. The scientific literature contains many examples of non-conformity score functions but there is an absence of studies examining their properties and effectiveness. In this paper, we give an overview of properties of non-conformity score functions. We give examples of non-conformity score functions in the existing literature and introduce original modifications. We introduce an original method of evaluating the prediction set sizes of conformal predictors and use it to provide a comparison between non-conformity score functions. We also examine efficacy of different non-conformity score functions for class-conditional conformal prediction in a setting with imbalanced classes.
\end{abstract}

\section{Introduction}

In machine learning for classification, an important aspect is that of model calibration. It is common practice to normalize classification output vectors and interpret the resulting numbers as probabilities. However, we expect true probabilities to satisfy certain criteria. Most commonly, in the frequentist interpretation, we expect that the probability corresponds to the frequency with which an outcome occurs over an infinite trial. For an uncalibrated machine learning model, this is not expected to hold. With conformal prediction, the model is allowed to predict a set of classes rather than one class. The algorithm with which such prediction sets are constructed allows calibration such that the \textit{a priori} probability of a prediction set of a conformal predictor containing the true class is at least as large as a pre-specified error rate. This is accomplished by assigning a value known as a non-conformity score (sometimes referred to as non-conformity measure) to every data point-label pair, and using the values of this score to determine whether a class should be included in the prediction set or not. For a formal mathematical theory of conformal prediction, see the second edition of \textcite{AlgorithmicLearningRandom2022}. For an informal description of conformal prediction, see the overview by \textcite{angelopoulosGentleIntroductionConformal2021}.

While any (non-conformity) score function produces valid prediction intervals using the conformal prediction algorithm, the ability of the score to accurately correspond to non-conformity matters greatly for the size and usefulness of these prediction intervals. In this article, we focus solely on score functions, giving an overview of some of the common choices in the literature and running benchmarks to gain insight into how to best assign non-conformity scores in a machine learning classification setting. Specifically, we aim to fill knowledge gaps with regard to the properties of non-conformity score functions, how they interact with models and model accuracy, and how well they handle class imbalance. In order to achieve this, we invent a metric for evaluating score functions, and use it to benchmark score functions applied on a variety of metric spaces and with a variety of distance metrics.

\subsection{Definition of conformal prediction}
\label{subsec:CPalg}

Conformal prediction (CP) is an algorithm that wraps around a predictive model such that it produces a set of classes rather than one class \parencite{shaferTutorialConformalPrediction2007a}. The algorithm in a classification context is described as follows:
\begin{enumerate}
    \item Given a predictive classification model and a calibration data set $\mathbb{C}$ not seen during training, denote by $X_i$ the i-th data point and $y_i$ its class (i.e., its label).
    \item Define a (non-conformity) score function
    \begin{equation} \label{score}
        s: ((X_i, y_i), \{\mathbb{C}\setminus(X_i,y_i)\}) \rightarrow \mathbb{R}, \quad s(X_i, y_i) = s_i.
    \end{equation}
    The score $s_i$ is a measurement of how strange an example's data $X_i$ is compared to examples in the calibration set belonging to its class $y_i$. The CP algorithm will work regardless of the function $s$, but the usefulness of the output varies greatly depending on its construction.
    \item Calibrate the conformal predictor by calculating $s_i$ for all the examples in $\mathbb{C}$.
    \item Choose an allowed error rate (sometimes called significance) $\alpha$.
    \item Run the conformal predictor on new data points $X_j$ with unknown class, not seen during training or calibration, by calculating $s_j^y = s(X_j, y)$ for each possible class $y$.
    \item Construct the prediction set $\mathcal{P}$ by including all classes with $s_j^c$ smaller than the $\frac{\lceil (|\mathbb{C}|+1)(1-\alpha) \rceil}{|\mathbb{C}|}$-th quantile of $s_i$ on $\mathbb{C}$. This quantile can be constructed per class (class-conditional or ``Mondrian" CP) or in total.
\end{enumerate}

\noindent Note that in (\ref{score}), we have made explicit the fact that statistics calculated on the entire calibration set can be used in the score function. However, if a score function does use values of the calibration set, we exclude the point itself to preserve exchangability (or alternatively, for transductive conformal predictors, include the test examples themselves during inference).

\subsection{Properties of conformal prediction}

The usefulness of CP comes from the mathematical guarantee that the underlying probability of coverage of the prediction set is larger than one minus the user-specified error rate $\alpha$ regardless of the underlying distribution of the data. The quantity $\frac{\lceil (|\mathbb{C}|+1)(1-\alpha) \rceil}{|\mathbb{C}|}$ in the CP algorithm is a finite-sample adjustment, so that this guarantee holds also for finite samples \parencite{angelopoulosGentleIntroductionConformal2021}. Importantly, this guarantee applies only to the underlying \textit{a priori} probability of coverage for a not-yet-calibrated conformal predictor, or, equivalently, for the average of all possible calibrations utilizing $|\mathbb{C}|$ elements of the combined calibration and test set. Once the calibration data split is determined, the underlying probability of coverage on any subsequent data with which the calibration data is exchangeable is beta-distributed \parencites{angelopoulosGentleIntroductionConformal2021}[Lemma 3]{vovk_conditional_2012}:

\begin{equation}
\mathbb{P} =   B(a, b), \text{for } 
a = |\mathbb{C}|+1-\lfloor{(|\mathbb{C}|+1)\alpha \rfloor}, b = \lfloor{(|\mathbb{C}|+1)\alpha \rfloor}.
\end{equation}

As $|\mathbb{C}|$ increases, this distribution becomes narrow around a point approximately equal to but not smaller than $1-\alpha$. Because this refers to underlying probability of marginal coverage, one cannot state for any specific prediction set the probability that it contains the true class, but the frequency of coverage on a large test set will approach $\mathbb{P}$, which approximates $1-\alpha$. It is worth stating clearly the fact that finite samples apply twice, first the finite sample of the calibration set which determines the underlying probability of coverage $\mathbb{P}$ for a specific conformal predictor (sampled from the beta distribution), then the finite sample of the test set which influences the variance of the observed coverage.

\subsection{The conformal domain}

Modern predictive classification models using deep learning will typically have a structure consisting of an encoder mapping the input into features, followed by processing of the features using e.g. transformers or other deep learning modules, followed (not always) by linear layers $L$ that map the processed features into a space with dimensionality equal to the number of classes being predicted. Finally, a softmax or related function to normalize this ``logit vector" into an output vector that is often interpreted as containing (uncalibrated) probabilities. If we combine the encoder and feature processing into a ``black box" function $\Box$, and denote by $f$ the predictive classification model with $n$ distinct classes, then

\begin{equation}
\label{domains}
f: X \xrightarrow{\Box} \mathbb{R}^k \xrightarrow{L} \mathbb{R}^n \xrightarrow{\sigma} \Delta^{n-1},
\end{equation}
with $k$ denoting the dimensionality of the feature space prior to the linear layers, $\Delta^n$ the standard n-simplex of probabilities, and $\sigma$ the softmax function which maps each value $x_j$ in a vector $v$ thusly:

\begin{equation}
    \label{softmax}
    \sigma_j(v) = \frac{e^{x_j}}{\sum_{j=1}^n{e^{x_j}}}.
\end{equation}

\noindent If there are indeed linear layers, then commonly we have $k \gg n$ and $L: \mathbb{R}^{k} \rightarrow \mathbb{R}^{k_1} \rightarrow \dots \rightarrow \mathbb{R}^{k_m} \rightarrow \mathbb{R}^{n}$, with $k \geq k_1 \geq \dots \geq k_m \geq n$. In this paper, we will use the following terminology: a (non-conformity) score function is a measurable function that takes inputs from points in a metric space $M$ along with their corresponding labels from a metric space $Y$ and returns a (non-conformity) score via
\begin{equation}
(M \times Y)^{|C|-1}\ \times (M \times Y) \rightarrow \mathbb{R} \quad \text{(calibration)},
\end{equation}
\begin{equation}
(M \times Y)^{|C|}\ \times (M \times Y) \rightarrow \mathbb{R} \quad \text{(inference)}.
\end{equation}
We refer to $M$ as the conformal domain, and $Y$ as the label space. It is common to not define score functions using the data points of the calibration space, in which case the non-conformity score function is a mapping $(M \times Y) \rightarrow \mathbb{R}$, however we will see that utilizing the mean calibration data point on the conformal domain constitutes a rather useful score function.

Most commonly, the conformal domain is the standard $(n-1)$-simplex of probabilities $\Delta^{n-1}$ obtained after applying softmax, but this is not a requirement. The primary usefulness of the simplex is that classes are representable as points in the space via their corresponding one-hot-encoded label vectors. In contrast, the logit space $\mathbb{R}^n$ would need to represent the label as a continuous one-dimensional set of points, due to the non-injectiveness of $\sigma$. The logit vectors do however still have a notion of rank identical to that of the simplex, in the sense that larger values are deemed more probable by the predictive model. In contrast again, the rank notion is lost when considering feature vectors. Still, any space along the chain in (\ref{domains}) can constitute the conformal domain, as long as the score function is chosen without direct utilization of labels or ranks.

\subsection{Evaluation metrics}
\label{sec:evaluation}

\begin{figure}
    \centering
    \includegraphics[width=0.90\linewidth]{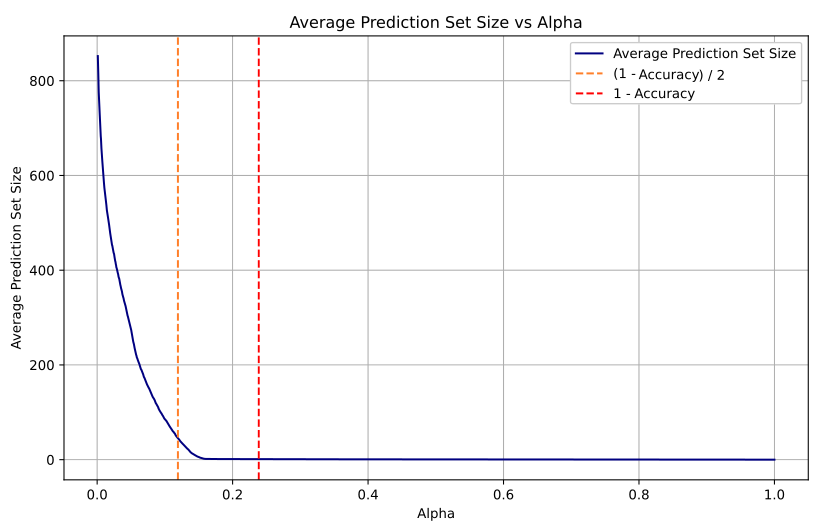}
    \caption{The prediction set size as a function of the error rate $\alpha$ for a conformal predictor predicting classes from the $1000$ classes of ImageNet, using a ResNet-50 classification model with $76.15\%$ accuracy. We plot also vertical lines where the error rate is equal to one minus the model accuracy and half of one minus the model accuracy.}
    \label{fig:size_over_alpha}
\end{figure}

Assuming a correct CP implementation and a reasonably sized calibration set, the only thing separating distinct CP algorithms with the same predictive model and conformal domain is the choice of score function. Method papers for CP will often evaluate their score function by reporting the coverage and the mean prediction set size
\begin{equation}
\label{meansetsize}
\widetilde{|\mathcal{P}|} = \frac{1}{|\mathcal{T}|}\sum_{i=1}^{|\mathcal{T}|}|\mathcal{P}_i|,
\end{equation}
where $\mathcal{T}$ is the test set not seen during model training nor CP calibration. However, any deviation from a coverage of $1-\alpha$ can only stem from stochasticity or incorrect implementation, and hence need not be reported as an evaluation metric. Mean prediction set size is however an important evaluation metric, where smaller is better. Worth noting is that prediction sets may be empty, and that this tends to only happen with notable frequency if the error rate is set higher than necessary: since a coverage of approximately $1-\alpha$ is guaranteed, a small error rate leaves very little or no room for empty sets. As the error rate is reduced, empty sets will become scarce. Hence, empty sets are not an issue for the metric (\ref{meansetsize}), lower is better despite them. For visualization, in Figure \ref{fig:size_over_alpha} we plot the quantity $\widetilde{|\mathcal{P}|}$ for a conformal predictor ran at different error rates on an underlying ResNet-50 classifier on the ImageNet dataset of $1000$ classes.

The evaluation metric (\ref{meansetsize}) is not sufficient to completely evaluate a CP algorithm. Indeed, if it were, the error rate would merely be set very high to incentivize small or empty sets. For this reason, evaluations of CP algorithms usually involve calculating $\widetilde{|\mathcal{P}|}$ at some different values of $\alpha$. The authors are not aware of any attempts at formulating evaluation metrics that produce a single number using both $\alpha$ and $\widetilde{|\mathcal{P}|}$, despite the obvious usefulness of such a metric. One suggestion is to compute
\begin{equation}
\label{integral}
I = \int_0^1{\widetilde{|\mathcal{P}|(\alpha)} d\alpha},
\end{equation}
i.e., the area under the curve in Figure \ref{fig:size_over_alpha}. This is not significantly more computationally intensive than computing $\widetilde{|\mathcal{P}|(\alpha)}$, since scores do not depend on $\alpha$ and thus the integral can be computed by retroactively constructing prediction intervals using pre-computed scores.

Another issue with evaluating conformal predictors in general is that the precision of the prediction sets will be influenced by properties of the underlying model, meaning comparing the quantity in (\ref{meansetsize}) across papers is usually not possible unless the model is exactly replicable. Though not a total solution to this issue, evaluating at an error rate that is computed from the accuracy of the model is a partial mitigation. For instance, if $A$ denotes model accuracy (the proportion of test examples for which the true class is given the highest probability), then the prediction set size at $\alpha_{1/2} = (1-A)/2$ may be a more objective measure of the conformal predictor than computing it at a fixed value of $\alpha$ across different models and datasets with varying accuracy. Similarly, one can pragmatically define
\begin{equation}
\label{integral2}
I_{1} = \frac{1}{1-A} \int_0^{1-A}{\widetilde{|\mathcal{P}|(\alpha)} d\alpha},
\end{equation}
to constrain the evaluation to the interesting values of $\alpha$, since, in practice it would be rare to want to use an $\alpha$ that allows a larger error than the model tends to produce. For instance, if the model's accuracy is $93\%$, one would compute the integral in $I_1$ with the error rate going from $0$ to $0.07$. This is the evaluation metric that will be used in place of (\ref{meansetsize}) in this paper. It is also conceivable to restrict the range of the integral from the left, since $\alpha=0$ is not a practical error rate and all conformal predictors will predict sets of nearly all classes near this value. However, for some applications a very small error rate is appropriate, and capturing in our evaluation metric information on how quickly the conformal predictor begins producing reasonably sized prediction sets as the error rate decreases is useful. Hence, while we are open to other integration bounds, in this paper we will estimate this integral numerically starting from $\alpha = 0.001$.

\section{Score functions}
\subsection{Label Distance}
\label{subsec:label_distance}
One very simple score function measures the distance between the prediction and the label. This makes sense to do when a model has been trained such that the outputs (on the $\Delta^{n-1}$-simplex) approximate basis vectors that are $1$ for the index representing the prediction and $0$ elsewhere. Denote by $d$ a general distance function, then the Label Distance score function is defined by
\begin{equation}
\label{eq:label_distance}
s_{label}: \Delta^{n-1} \rightarrow \mathbb{R}, \quad s_{label} (X_i, y_i) = d(f(X_i), y_i).
\end{equation}

The distance $d$ for this specific score function is usually an $L^p$-norm. This score function in a regression context is introduced in early texts on conformal prediction \parencite{shaferTutorialConformalPrediction2007a} and in \textcite{sadinleLeastAmbiguousSetValued2016} it is used in classification. This score function can only be computed on $\Delta^{n-1}$, since the label must be representable as a point in the conformal domain.

\subsection{Margin Distance}

Another simple approach measures the distance not to the label vector, but to the decision boundary between a correct and an incorrect prediction. If the current highest probability output is not the true class, this is the shortest distance to a point in the conformal domain that does correspond to a correct prediction; if it is already the true class then it is the negative distance to the nearest point in the conformal domain that corresponds to an incorrect prediction. Let $\hat{\pi}^{(y)}$ be the predicted value (logit or probability) belonging to the true class. The decision boundaries between correct and incorrect predictions are all points that have $\hat{\pi}^{(y)} = \hat{\pi}^{(x)}, x \neq y$. In binary classification on the probability 1-simplex this is trivially the point $(0.5, 0.5)$, whereas in binary classification in two-dimensional logit space this is all points along the line $x=y$. In the general multi-class setting there are many boundaries and the Margin Distance score function measures the distance to the nearest of them. This requires some care, since what is ``nearest" depends on the distance metric. Let the closest point on any decision boundary as measured by the distance function $d$ be the point $b_d$. Then
\begin{equation}
s_{margin}: \mathbb{R}^n \rightarrow \mathbb{R}, \quad s_{margin} (X_i, y_i) = d(f(X_i), b_d(y_i)).
\end{equation}
We work it out explicitly for Euclidean distance and cosine distance. Let $\hat{\pi}^{(z)}, z \neq y$ be the highest predicted probability not belonging to the true class. For both these distance metrics, the nearest decision boundary $b$ is that with the elements corresponding to classes $y,z$ equal. Let this value be $a$. The projection of $\hat{\pi}$ onto the decision boundary is retrieved from minimizing $(\hat{\pi}^{(y)}-a)^2+(\hat{\pi}^{(z)}-a)^2$ while letting all other elements be the same as in $\hat{\pi}$. This gives
\begin{equation}
a = b^{(y)} = b^{(z)} = (\hat{\pi}^{(y)}+\hat{\pi}^{(z)})/2,
\end{equation}
so we have an explicit definition of the nearest boundary point. The Euclidean distance is the length of the difference vector between $\hat{\pi}$ and $b$. The Margin Distance, with signs defined as described above, is
\begin{equation}
\text{sgn}(\hat{\pi}^{(z)}-\hat{\pi}^{(y)}) || \ [ \frac{|\hat{\pi}^{(y)}-\hat{\pi}^{(z)}|}{2}, \frac{|\hat{\pi}^{(y)}-\hat{\pi}^{(z)}|}{2}] \ ||
\end{equation}
which equivalently for CP can be replaced with
\begin{equation}
\hat{\pi}^{(z)}-\hat{\pi}^{(y)},
\end{equation}
since in both cases all scores will be ordered based on the value of $\hat{\pi}^{(z)}-\hat{\pi}^{(y)}$. The cosine distance (one minus cosine similarity) is given by
\begin{equation}
1-\frac{b \cdot \hat{\pi}}{||b||||\hat{\pi}||},
\end{equation}
but note that
\begin{align*}
    b \cdot \hat{\pi}
        &= \pi^{(y)}\left(\frac{\pi^{(y)} + \pi^{(z)}}{2}\right)
         + \pi^{(z)}\left(\frac{\pi^{(y)} + \pi^{(z)}}{2}\right)
         + \sum_{x \ne y,z} \bigl(\hat{\pi}^{(x)}\bigr)^2 \\
        &= 2 \left( \frac{\pi^{(y)} + \pi^{(z)}}{2} \right)^2
         + \sum_{x \ne y,z} \bigl(\hat{\pi}^{(x)}\bigr)^2 = \lVert b \rVert^2,
\end{align*}
so that
\begin{equation}
1-\frac{b \cdot \hat{\pi}}{||b||||\hat{\pi}||} = 1-\frac{||b||}{||\hat{\pi}||}.
\end{equation}
Naturally we let the score be negative when the highest probability prediction is already the true class, since then a larger distance to the boundary means the model is deeper in the region of correct predictions and hence more conforming. This score function using Euclidean distance was introduced by \textcite{lofstromBiasReductionConditional2015a}, where it was applied on $\Delta^{n-1}$. Applying the Margin Distance score function with cosine distance as a distance metric is, as far as the authors know, novel.

\subsection{Mean Distance}
A flexible alternative to the Label Distance in \ref{subsec:label_distance} is to measure the distance to the mean point in the calibration set belonging to the same class:
\begin{equation}
\label{eq:mean_distance}
s_{\text{mean}}: (X_i, y_i, \{\mathbb{C}\setminus(X_i,y_i)\}) = d(f(X_i), \text{mean}(\{X_j \in \{\mathbb{C}\setminus(X_i,y_i)\} \ \forall \ y_j=y_i \})).
\end{equation}

This was first done by \textcite{shaferTutorialConformalPrediction2007a}. Unlike the previous score functions mentioned, this is just as easy to calculate in any space in (\ref{domains}), including in feature space, as it is on the logit or output space. This can be generalized into KMeans (or KMedians) approaches, which helps if the datapoints in the conformal domain are divided into multiple clusters. However, in order to preserve exchangability, the KMeans clusters need to be re-computed for each calibration example, which requires heavy computation.

\subsection{Nearest neighbor score functions}

Using a nearest neighbors algorithm as a score function was also introduced by \textcite{shaferTutorialConformalPrediction2007a}, where the score was computed as the distance to the nearest point in the calibration set of the same class divided by the distance to the nearest point in the calibration set of a different class. Naturally, many alternative versions to this which utilize a K-nearest neighbors algorithm can be used as well. Like the Mean Distance score functions, these can be computed in any space, as they do not require representing the label in the conformal domain.

\subsection{APS, RAPS and SAPS}

In a 2020 paper titled \textit{Classification with Valid and Adaptive Coverage}, \textcite{romanoClassificationValidAdaptive2020a} introduced a method which has been commonly referred to as Adaptive Prediction Sets, or ``APS". The proposed algorithm differs from the CP algorithm in section \ref{subsec:CPalg}, but we show their equivalence in Appendix \ref{sec:appendixA}. Their score function is equal to the the cumulative sum of ordered outputs from the classification model, from the largest until the one corresponding to the true class, the last of which is weighted by a random uniform variable. This method was extended later the same year by \textcite{angelopoulosUncertaintySetsImage2020a}, where they added a regularization term to the score function. This method is known as Regularized Adaptive Prediction Sets, or ``RAPS". The full score function including the regularization term is
\begin{equation}
\label{eq:raps}
s_{\text{RAPS}}(X,y) = \sum_{i, \hat{y}^i>\hat{y}^c}\hat{y}^i + u \cdot \hat{y}^c + \lambda [(r(\hat{y},y)-k_{\text{reg}})]^+,
\end{equation}
where $\hat{y}$ is the output vector of predicted probabilities, $u \sim U(0,1)$ is a random variable from a uniform distribution, $r(\hat{y},y)$ is the rank of the label (e.g., if the true class has the fourth highest predicted probability, the rank is $r(\hat{y},y) = 4$), and $\lambda$ and $k_{\text{reg}}$ are regularization parameters. The plus sign indicates that the regularization term is only added to the score if it is positive. Intuitively, the regularization term encourages smaller set sizes by increasing the score for all classes whose rank is greater than the cutoff $k_{reg}$, which in the paper on RAPS reports impressive improvements on $\widetilde{P(\alpha)}$ over APS.\\

\noindent Another paper by \textcite{huangConformalPredictionDeep2023a} reports slight improvements on $\widetilde{P(\alpha)}$ over RAPS with minor modifications to the score function. It disregards all predicted probabilities other than the highest. They write:
\begin{equation}
\label{eq:saps}
s_{\text{SAPS}}(X,y) = \begin{cases} 
u \cdot \sup\{{\hat{y}^i}\}, \quad r(\hat{y},y)=1 \\\sup\{\hat{y}^i\} + \lambda (r(\hat{y},y)-2+u), \quad r(\hat{y},y) \geq 2.
\end{cases}
\end{equation}
\noindent This is in many ways a simpler score function than (\ref{eq:raps}). The motivation is that the predicted probabilities of the model are noisy and that low probability predictions are not necessarily of any value. It is purposefully constructed such that it yields the same score as APS when the rank of the true (calibration) or test (prediction) class is the highest probability prediction.

These paper use the probability simplex as its conformal domain but this score function can be calculated in the logit space, though not in feature space where the notion of rank is lost. If the output probability term is removed from the SAPS score function, it yields the same results in output and logit space, since then it is purely rank-based.

The prediction set size of these algorithms on the probability simplex can be related to top-k accuracy in several interesting ways, most importantly being that RAPS can always do better (smaller prediction sets) than the top-k accuracy with $k$ such that coverage is achieved \parencite{angelopoulosUncertaintySetsImage2020a}. For more similarly interesting results, see the papers by \textcites{angelopoulosUncertaintySetsImage2020a, huangConformalPredictionDeep2023a}

\subsubsection{The randomized term is not required for exact coverage}
The argument presented by both \textcites{romanoClassificationValidAdaptive2020a, angelopoulosUncertaintySetsImage2020a} for inclusion of the randomized variable $u$, is that this is required to achieve exact coverage. The argument is that if one always added terms until one exceeds the calibrated threshold one would end up with higher than intended coverage, while if one always stopped one term prior to exceeding the threshold one would end up with lower than intended coverage, and that a medium between the two is necessary for exact coverage. However, for calibrated conformal predictors, the CP algorithm provided in \ref{subsec:CPalg} is guaranteed to provide coverage with probability pre-specified by the error rate, regardless if one includes the randomized term or not. The purpose of the randomized term in its first mention, by \textcite{romanoClassificationValidAdaptive2020a}, was in the context of a hypothetical perfectly calibrated model in which $1-\alpha$ could be used as a threshold directly. Then, the randomized term was required since no calibration took place, and exceeding the threshold during inference would lead to high coverage if not for the random discarding. In Appendix \ref{sec:appendixA} we show the equivalence of this uncalibrated algorithm with the conformal prediction algorithm \ref{subsec:CPalg} where the random term is included in the score function. However, since this is now a score function in a CP algorithm, the random term can be removed or replaced with any constant value. For instance, we have
\begin{equation}
    s_{\text{RAPS}}(X,y,u=0) =  \sum_{i, \hat{y}^i>\hat{y}^c}\hat{y}^i + \lambda [(r(\hat{y},y)-k_{\text{reg}})]^+,
\end{equation}
\begin{equation}
    s_{\text{RAPS}}(X,y,u=1) =  \sum_{i, \hat{y}^i \geq \hat{y}^c}\hat{y}^i + \lambda [(r(\hat{y},y)-k_{\text{reg}})]^+,
\end{equation}
both of which have the coverage guarantee if ties are disregarded. With regard to ties, the former of the above equations (for $u=0$) will give a score of $s=0$ whenever the rank is $1$, which means that any model with top-1 accuracy greater than $1-\alpha$ will get coverage equal to the top-1 accuracy as opposed to coverage equal to $1-\alpha$, since the calibrated threshold will be $0$. For this reason it is pragmatic to let $U>0$ merely as a tie-break factor.

Note that letting $U$ be small in the APS score function partially mitigates the need for regularization, since it (similarly to regularization with $k_\text{reg} = 1$) separates the examples where the true class had the highest predicted probability, so that the score function will be close to $0$, from the cases where this is not true. For this reason, smaller values for $U$ may greatly improve the performance of APS while not having significant impact on RAPS or SAPS.

\subsection{Gradient Distance (Feature and Fast Feature Conformal Prediction)}
Feature CP, in this paper referred to as the Gradient score function, was the name of a method introduced by \textcite{tengPredictiveInferenceFeature2022a} that is explicitly intended to use feature space as its conformal domain. This is, however, not a requirement for the score function introduced in the paper, which similarly to other distance-based score functions can be used in any domain. A modification to the score function was done by \textcite{tangPredictiveInferenceFast2024}, providing significant computational speed-up. The theoretical score function for a feature vector $\hat{v}$ introduced in these two papers is:
\begin{equation}
\label{eq:featurecp}
s = \inf_{v: g(v)=Y} ||v-\hat{v}||,
\end{equation}
where $v$ is any point in the feature space, so that $v: g(v) = Y$ are all points in feature space which would map to exactly the correct output if one applied ``the rest" ($g$) of the neural network to that feature. This is conceptually very similar to the label score function, except it is applied in feature space. Computing (\ref{eq:featurecp}) exactly is not possible, as it would require determining the downstream output for all points in the feature space. \textcite{tengPredictiveInferenceFeature2022a} approximated this numerically by performing gradient descent on the feature vector for a fixed number of iterations and recording the difference between the start and end points. In the paper by \textcite{tangPredictiveInferenceFast2024} it is instead approximated by the quantity
\begin{equation}
\label{eq:fastfeaturecp}
s = ||v-\hat{v}|| \approx \frac{||Y-g(\hat{v})||}{||\nabla g(\hat{v})||},
\end{equation}
which is a first order Taylor approximation. The main benefit of the Fast Feature CP, in this paper referred to as the Fast Gradient score function, is that it mitigates the computational time required to compute gradient descent for several iterations for each test example and label combination. However, the (fairly limited) benchmarking done by \textcite{tangPredictiveInferenceFast2024} does also report minor improvements to prediction set size over the iterative feature CP method on certain tests, suggesting there may be qualitative performance gains as well.

\section{Evaluation of score functions}
In this section, we aim to produce some evaluations to compare the score functions described in the previous section, in a systematic and reproducible fashion.

\subsection{Technical implementation}
The classification datasets used are ImageNet, CIFAR10, and CIFAR100. The primary model is Resnet50 \parencite{heDeepResidualLearning2015}, pre-trained on ImageNet and (except for ImageNet classification itself) fine-tuned for exactly 50 epochs on the corresponding dataset without early stopping. We compute $I_1$ as in (\ref{integral2}) in a non-class-conditional setting. We then report results on three auxiliary models: ResNet18, EfficientNet-B0 \parencite{tanEfficientNetRethinkingModel2020}, and ViT-B16 \parencite{transformer}, the first of which has similar architecture but worse accuracy than ResNet50 and the latter two which have similar or slightly higher accuracy but substantially different architecture to ResNet-50, in order to see whether model architecture or model accuracy has an effect on which methods perform the best. For a more detailed description of model training and evaluation, see appendix \ref{sec:appendixC}.

For CIFAR10 and CIFAR100 we let $20,000$ examples be unseen during model training, use $10,000$ for calibration of the CP model, and $10,000$ for testing the CP model (the results which we report). For ImageNet, we use the entire ImageNet validation set from 2012 (ILSVRC2012) of $50,000$ images held out, resulting in $40,000$ calibration examples and $10,000$ test examples. CIFAR10 has $10$ classes, CIFAR100 has $100$ classes and ImageNet has $1000$ classes.

\subsection{Computing $I_1$}

We computed the value $I_1$ in (\ref{integral2}) computing the mean prediction set size for $50$ values of $\alpha$, from $0.001$ to $1-A_1$ and approximating using the trapezoidal rule for numerical integration. The same randomization of calibration and test set is used for each run of CP in the computation of $I_1$. The total procedure is then repeated over $100$ different calibration-test randomizations. We report the median over these $100$ runs. For ResNet50, all score functions were evaluated in all conformal domains that are applicable for that score, between probability space, logit space, and feature space. For each score function, we report only the best-performing version, e.g. if Margin Distance performs better in logit space than in probability space then only the former is reported. The results are presented in Table \ref{table1}. ``Feature space" is here the space of the final feature vector prior to the output (logit) layer. For ResNet50 it has 2048 dimensions, for ResNet-18 it has 512 dimensions, for EfficientNet-B0 it has 1240 dimensions and for ViT-B16 it has 768 dimensions. The full results showing all tested versions of all score functions for all models is shown in Supplementary Table 1 in Appendix \ref{sec:appendixA}.

\newpage

\begin{table}[]
\centering
\makebox[\textwidth][c]
{%
\begin{tabular}{@{}llllll@{}}
\toprule
\multicolumn{1}{c}{Dataset} &
  \multicolumn{1}{c}{\makecell[c]{Model \\ Accuracy}} &
  \multicolumn{1}{c}{Score function} &
  \multicolumn{1}{c}{{\makecell[c]{Conformal domain}}} &
  \multicolumn{1}{c}{\makecell[c]{Distance\\metric}} &
  \multicolumn{1}{c}{$I_1$ ($Q_1$--$Q_3$)} \\ \midrule
  ImageNet & 76.15\% & Label Distance & Probability space & Cosine Distance & 8.46 (8.37--8.59)\\
 &  & Margin Distance & Logit space & Cosine Distance & 8.12 (8.03--8.24) \\
  &  & Mean Distance & Probability space       & Cosine Distance & 8.53 (8.39--8.61) \\
 &  & APS (U=0.001)   & Probability space & N/A                &  15.5 (15.2--15.8)\\
  &  & RAPS ($\lambda=0.2, k=1$, U=0.001) & Probability space & N/A & 9.92 (9.82--10.0)\\ 
 &  & SAPS ($\lambda=0.2$, U=0.001) & Probability space & N/A & 9.60 (9.49--9.71) \\ 
  &  & \textbf{Fast Gradient} & \textbf{Feature space} & \textbf{Euclidean Distance} & \textbf{7.64 (7.54--7.76)}\footnotemark\\
\midrule
CIFAR100 & 84.73\% & Label Distance  & Probability space & Cosine Distance & 5.57 (5.44--5.71)\\
 &  & Margin Distance & Logit space & Cosine Distance & 5.35 (5.20--5.48) \\
  &  & \textbf{Mean Distance} & \textbf{Feature space}       & \textbf{Cosine Distance} & \textbf{2.78 (2.73--2.87)} \\
 &  & APS (U=0.001)   & Probability space & N/A                &  6.71 (6.51--6.84)\\
  &  & RAPS ($\lambda=0.2, k=1$, U=0.001) & Probability space & N/A & 4.92 (4.77--5.07)\\ 
 &  & SAPS ($\lambda=0.2$, U=0.001) & Probability space & N/A & 4.75 (4.63--4.90) \\ 
 &  & Gradient (lr=0.1, 100 steps) & Feature space & Euclidean Distance & 4.65 (4.50--4.77) \\ 
 &  & Fast Gradient & Feature space & Euclidean Distance & 4.25 (4.15--4.39) \\ 
\midrule
CIFAR10 & 96.93\% & Label Distance  & Probability space & Cosine Distance & 1.14 (1.13--1.16)\\
 &  & Margin Distance & Logit space       & Cosine Distance & 1.11 (1.10--1.12) \\
  &  & \textbf{Mean Distance} & \textbf{Logit space}       & \textbf{Euclidean Distance} & \textbf{1.08 (1.07--1.09)}\\
 &  & APS (U=0.001)   & Probability space & N/A                & 1.19 (1.16--1.22) \\
  &  & RAPS ($\lambda=0.2, k=1$, U=0.001) & Probability space & N/A & 1.35 (1.32--1.38) \\ 
 &  & SAPS ($\lambda=0.2$, U=0.001) & Probability space & N/A & 1.26 (1.23--1.29) \\ 
 &  & Gradient (lr=0.1, 100 steps) & Feature space & Euclidean Distance & 1.24 (1.20--1.27) \\ 
 &  & Fast Gradient & Feature space & Euclidean Distance & 1.12 (1.11--1.14) \\ 
 \bottomrule
\end{tabular}
}
\caption{Values of $I_1$, defined in (\ref{integral2}), for different score functions, conformal domains, and distance metrics on the ImageNet, CIFAR100, and CIFAR10 datasets, with a ResNet-50 model. Median, first quartile and third quartile across $100$ runs. Lower is better. Best score function for each dataset is presented in bold. For a version with all versions of all score functions see \ref{totaltable}.}
\label{table1}
\end{table}

\footnotetext{In version v1 of this pre-print on arXiv, a bug caused the score of the Fast Gradient score function to be much higher in all evaluations.}

\FloatBarrier

For the remaining models, the version that was best-performing on ResNet50 is reported, along with whichever versions outperformed those that were best-performing on ResNet50 for the respective score functions. The results are in Table \ref{table2}.

\newpage

\setlength{\LTleft}{-3.5cm}  

\begin{longtable}{@{}lllllll@{}}
\\
\toprule
\multicolumn{1}{c}{Model} & \multicolumn{1}{c}{Dataset} & \multicolumn{1}{c}{\makecell[c]{Model \\ Accuracy}} & \multicolumn{1}{c}{Score function} & \multicolumn{1}{c}{\makecell[c]{Conformal\\domain}} &   \multicolumn{1}{c}{\makecell[c]{Distance\\metric}} & \multicolumn{1}{c}{$I_1$ ($Q_1$--$Q_3$)} \\ 
\midrule
\endfirsthead

\multicolumn{6}{c}{\tablename\ \thetable\ -- \textit{Continued from previous page}} \\
\toprule
\multicolumn{1}{c}{Model} & \multicolumn{1}{c}{Dataset} & \multicolumn{1}{c}{\makecell[c]{Model \\ Accuracy}} & \multicolumn{1}{c}{Score function} & \multicolumn{1}{c}{\makecell[c]{Conformal\\domain}} &   \multicolumn{1}{c}{\makecell[c]{Distance\\metric}} & \multicolumn{1}{c}{$I_1$ ($Q_1$--$Q_3$)} \\ 
\midrule
\endhead

\midrule
\multicolumn{6}{r}{\textit{Continued on next page}} \\
\endfoot

\bottomrule
\endlastfoot
  ResNet18 & ImageNet & 69.76\% & Label Distance & Probabilities & Cosine & 10.6 (10.5--10.8)\\
 & &  & Margin Distance & Logits & Cosine & 10.4 (10.2--10.5)\\
 &  &  & Mean Distance & Probabilities       & Cosine & 10.9 (10.8--11.0) \\
 & &  & APS (U=0.001)   & Probabilities & N/A                &  16.8 (16.6--17.0)\\
 & &  & RAPS ($\lambda=0.2, k=1$, U=0.001) & Probabilities & N/A & 14.1 (14.0--14.2)\\ 
& &  & SAPS ($\lambda=0.2$, U=0.001) & Probabilities & N/A & 13.7 (13.6--13.8) \\
 & &  & \textbf{Fast Gradient} & \textbf{Features} & \textbf{Euclidean} & \textbf{9.70 (9.61--9.80)}\\ 
\midrule
 & CIFAR100 & 81.50\% & Label Distance  & Probabilities & Cosine & 5.07 (4.96--5.18)\\
 & &  & Margin Distance & Logits       & Cosine & 5.04 (4.90--5.13)\\
 & & & \textbf{Mean Distance} & \textbf{Probabilities}       & \textbf{Cosine} & \textbf{3.12 (3.05--3.18)} \\
 &  &  & \textit{Mean Distance} & \textit{Features}       & \textit{Cosine} & \textit{3.48 (3.43--3.54)} \\
 & &  & APS (U=0.001)   & Probabilities & N/A                &  6.50 (6.39--6.63)\\
 &  &  & RAPS ($\lambda=0.2, k=1$, U=0.001) & Probabilities & N/A & 6.01 (5.87--6.17)\\ 
 & &  & SAPS ($\lambda=0.2$, U=0.001) & Probabilities & N/A & 5.81 (5.68--5.94)\\ 
 & &  & Gradient (lr=0.1, 100 steps) & Features & Euclidean & 4.20 (4.10--4.31) \\ 
 & &  & Fast Gradient & Features & Euclidean & 4.22 (4.13--4.35)\\ 
\midrule
 & CIFAR10 & 96.00\% & Label Distance  & Probabilities & Cosine & 1.18 (1.17--1.19)\\
 & &  & Margin Distance & Logits       & Cosine & 1.17 (1.15--1.18) \\
 &  &  & \textbf{Mean Distance} & \textbf{Logits}       & \textbf{Euclidean} & \textbf{1.12 (1.11--1.14)}\\
 & &  & APS (U=0.001)   & Probabilities & N/A                & 1.19 (1.17--1.21) \\
 &  &  & RAPS ($\lambda=0.2, k=1$, U=0.001) & Probabilities & N/A & 1.32 (1.30--1.35) \\ 
 & &  & SAPS ($\lambda=0.2$, U=0.001) & Probabilities & N/A & 1.23 (1.22--1.26) \\ 
 & &  & Gradient (lr=0.1, 100 steps) & Features & Euclidean & 1.16 (1.15--1.17) \\ 
 & &  & Fast Gradient & Features & Euclidean & 1.14 (1.13--1.16) \\ 
\midrule
EfficientNet-B0 & ImageNet & 77.67\% & Label Distance & Probabilities & Cosine & 10.7 (10.5--10.9)\\
& & & Margin Distance & Logits & Euclidean & 11.3 (11.1--11.5)\\
 &  &  & \textbf{Mean Distance} & \textbf{Probabilities} & \textbf{Cosine} & \textbf{7.36 (7.20--7.46)} \\
 & & & Mean Distance & Logits & Cosine & 10.4 (10.2--10.6)\\
 & &  & APS (U=0.001)   & Probabilities & N/A                &  26.2 (25.8--26.5)\\
 & &  & RAPS ($\lambda=0.2, k=1$, U=0.001) & Probabilities & N/A & 11.9 (11.7--12.2)\\ 
& &  & SAPS ($\lambda=0.2$, U=0.001) & Probabilities & N/A & 11.6 (11.4--11.8) \\ 
& &  & \textit{Fast Gradient} & \textit{Features} & \textit{Euclidean} & \textit{9.68 (9.62--9.80)} \\ 
\midrule
 & CIFAR100 & 85.22\% & Label Distance  & Probabilities & Cosine & 3.79 (3.69--3.88)\\
 & &  & Margin Distance & Logits       & Cosine & 3.93 (3.84--4.02) \\
 &  &  & \textbf{Mean Distance} & \textbf{Features}       & \textbf{Cosine} & \textbf{2.31 (2.26--2.33)} \\
 & &  & APS (U=0.001)   & Probabilities & N/A                &  5.80 (5.67--5.94)\\
 &  &  & RAPS ($\lambda=0.2, k=1$, U=0.001) & Probabilities & N/A & 4.24 (4.11--4.37)\\ 
 & &  & SAPS ($\lambda=0.2$, U=0.001) & Probabilities & N/A & 4.05 (3.91--4.18) \\ 
 & &  & Gradient (lr=0.1, 100 steps) & Features & Euclidean & 3.26 (3.21--3.36) \\ 
 & &  & Fast Gradient & Features & Euclidean & 3.27 (3.13--3.35) \\ 
\midrule
 & CIFAR10 & 97.09\% & Label Distance  & Probabilities & Cosine & 1.13 (1.11--1.16)\\
 & &  & Margin Distance & Logits & Cosine & 1.11 (1.10--1.12) \\
 &  &  & \textbf{Mean Distance} & \textbf{Probabilities} & \textbf{Cosine} & \textbf{1.07 (1.07--1.08)}\\
 &  &  & \textit{Mean Distance} & \textit{Logits} & \textit{Euclidean} & \textit{1.14 (1.12--1.16)}\\
 & & & Mean Distance & Features & Cosine & 1.13 (1.12--1.14) \\
 & &  & APS (U=0.001)   & Probabilities & N/A                & 1.24 (1.20--1.26) \\
 &  &  & RAPS ($\lambda=0.2, k=1$, U=0.001) & Probabilities & N/A & 1.37 (1.33--1.40) \\ 
 & &  & SAPS ($\lambda=0.2$, U=0.001) & Probabilities & N/A & 1.28 (1.24--1.30) \\ 
 & &  & Gradient (lr=0.1, 100 steps) & Features & Euclidean & 1.13 (1.10--1.15) \\ 
 & &  & Fast Gradient & Features & Euclidean & 1.11 (1.10--1.13) \\ 
\midrule
ViT-B16 & ImageNet & 81.07\% & Label Distance & Probabilities & Cosine & 8.93 (8.66--9.10)\\
& &  & Margin Distance & Logits       & Euclidean & 9.54 (9.30--9.77)\\
 &  &  & \textbf{Mean Distance} & \textbf{Probabilities} & \textbf{Cosine} & \textbf{6.88 (6.74--7.00)} \\
 & &  & APS (U=0.001)   & Probabilities & N/A                &  25.8 (25.4--26.2)\\
 & &  & RAPS ($\lambda=0.2, k=1$, U=0.001) & Probabilities & N/A & 11.4 (11.2--11.6)\\ 
& &  & SAPS ($\lambda=0.2$, U=0.001) & Probabilities & N/A & 11.1 (11.0--11.3) \\ 
 & &  & \textit{Fast Gradient} & \textit{Features} & \textit{Euclidean} & \textit{8.39 (8.10--8.91)} \\ 
\midrule
 & CIFAR100 & 86.74\% & Label Distance  & Probabilities & Cosine & 6.08 (5.91--6.28)\\
 & &  & Margin Distance & Logits       & Cosine & 5.89 (5.75--6.06) \\
  & &  & \textbf{Mean Distance} & \textbf{Probabilities}       & \textbf{Cosine} & \textbf{3.45 (3.32--3.55)} \\
  & &  & \textit{Mean Distance} & \textit{Features}       & \textit{Cosine} & \textit{5.35 (5.14--5.58)} \\
  & &  & Mean Distance & Logits       & Cosine & 3.63 (3.50--3.77) \\
 & &  & APS (U=0.001)   & Probabilities & N/A                &  7.26 (7.00--7.51)\\
 &  &  & RAPS ($\lambda=0.2, k=1$, U=0.001) & Probabilities & N/A & 6.16 (5.94--6.36)\\ 
 & &  & SAPS ($\lambda=0.2$, U=0.001) & Probabilities & N/A & 6.00 (5.82--6.23) \\ 
 & &  & Gradient (lr=0.1, 100 steps) & Features & Euclidean & 5.87 (5.61--6.05) \\ 
 & &  & Fast Gradient & Features & Euclidean & 5.14 (4.97--5.32) \\ 
\midrule
 & CIFAR10 & 97.66\% & Label Distance  & Probabilities & Cosine & 1.20 (1.15--1.23)\\
 & &  & Margin Distance & Logits & Cosine & 1.09 (1.07--1.12) \\
 & & & Mean Distance & Probabilities & Cosine & 1.11 (1.10--1.12) \\
 &  &  & \textit{Mean Distance} & \textit{Logits} & \textit{Euclidean} & \textit{1.14 (1.11--1.16)}\\
 &  &  & \textbf{Mean Distance} & \textbf{Logits} & \textbf{Cosine} & \textbf{1.08 (1.06--1.13)}\\
 & &  & APS (U=0.001)   & Probabilities & N/A                & 1.28 (1.24--1.31) \\
 &  &  & RAPS ($\lambda=0.2, k=1$, U=0.001) & Probabilities & N/A & 1.36 (1.32--1.40) \\ 
 & &  & SAPS ($\lambda=0.2$, U=0.001) & Probabilities & N/A & 1.28 (1.25--1.32) \\ 
 & &  & Gradient (lr=0.1, 100 steps) & Features & Euclidean & 1.31 (1.25--1.34) \\ 
 & &  & Fast Gradient & Features & Euclidean & 1.17 (1.15--1.21)\\
 \bottomrule
\caption{Same as Table \ref{table1} but for ResNet-18, EfficientNet-B0, and ViT-B16. All versions of score functions that performed best for ResNet50 are reported, as well as all versions which outperformed those that performed best for ResNet50. Median, first quartile and third quartile across $100$ runs. Best score function for each dataset is presented in bold. Italic indicates a score function that was not best but that was best for ResNet50.}
\label{table2}
\end{longtable}

\FloatBarrier

\subsection{Computing prevalence in a class-conditional setting with imbalanced data}

Using ResNet50 on CIFAR10 and CIFAR100, we tested how some of the best-performing score functions from the previous section performed in a class-conditional setting where certain classes had been reduced to $10\%$ of their original prevalence in both model training and CP calibration and testing. In particular, we tested with which frequency that the minority classes, on average, appeared in the prediction sets. For instance, a prevalence of $10\%$ implies that \textit{each} minority class was prevalent in $10\%$ of the prediction sets, on average. These tests used a constant error rate of $\alpha = 0.10$, a target that should be easy on the CIFAR10 data where model accuracy is $95.36\%$, and difficult on the CIFAR100 data where the model accuracy is $81.16\%$.

\begin{table}[]
\centering
\makebox[\textwidth][c]
{%
\begin{tabular}{@{}llllll@{}}
\toprule
\multicolumn{1}{c}{Dataset} &
  \multicolumn{1}{c}{\makecell[c]{Model \\ Accuracy}} &
  \multicolumn{1}{c}{Score function} &
  \multicolumn{1}{c}{Conformal domain} &
  \multicolumn{1}{c}{\makecell[c]{Distance\\metric}} &
  \multicolumn{1}{c}{\makecell[c]{Prevalence (\%)\\(true / expected)}}\\ \midrule
CIFAR100 & 81.16\% & Label Distance  & Probabilities & Cosine & 27.1 / 0.105\\
&  & Margin Distance & Logits       & Euclidean & 27.8 / 0.105 \\
 &  & Margin Distance & Logits & Cosine & 20.1 / 0.105 \\
 &  & Mean Distance & Logits       & Euclidean & 19.1 / 0.105\\
  &  & \textbf{Mean Distance} & \textbf{Logits}       & \textbf{Cosine} & \textbf{6.79 / 0.105} \\
 &  & Mean Distance & Features  & Euclidean & 16.2 / 0.105\\
  &  & Mean Distance & Features & Cosine & 7.82 / 0.105 \\
 &  & APS (U=0.001)   & Probabilities & N/A                &  24.4 / 0.105\\
  &  & RAPS ($\lambda=0.2, k=1$, U=0.001) & Probabilities & N/A & 12.6 / 0.105\\ 
 &  & SAPS ($\lambda=0.2$, U=0.001) & Probabilities & N/A & 12.0 / 0.105\\ 
 &  & Fast Gradient & Features & Euclidean & 8.42 / 0.105 \\
\midrule
CIFAR10 & 95.36\% & Label Distance  & Probabilities & Cosine & 1.50 / 1.08\\
&  & Margin Distance & Logits       & Euclidean & 1.55 / 1.08 \\
 &  & Margin Distance & Logits       & Cosine & 1.32 / 1.08 \\
  &  & Mean Distance & Logits       & Euclidean & 1.29 / 1.08\\
  &  & \textbf{Mean Distance} & \textbf{Logits} & \textbf{Cosine} & \textbf{1.22 / 1.08}\\
  &  & Mean Distance & Features       & Euclidean & 1.25 / 1.08\\
&  & \textbf{Mean Distance} & \textbf{Features}       & \textbf{Cosine} & \textbf{1.22 / 1.08} \\
 &  & APS (U=0.001)   & Probabilities & N/A                & 1.39 / 1.08 \\
  &  & RAPS ($\lambda=0.2, k=1$, U=0.001) & Probabilities & N/A & 1.25 / 1.08 \\ 
 &  & SAPS ($\lambda=0.2$, U=0.001) & Probabilities & N/A & 1.37 / 1.08 \\
 &  & Fast Gradient & Features & Euclidean & 1.31 / 1.08 \\
 \bottomrule
\end{tabular}
}
\caption{Values of the mean prevalence of minority classes for CIFAR100, where $10$ out of $100$ classes are minority, and CIFAR10, where $1$ out of $10$ classes is minority. These results are using class-conditional CP with some select score functions, conformal domains, and distance metrics on the CIFAR10 and CIFAR100 datasets, with a Resnet50 model. ``Expected" refers to the true mean prevalence of the class, i.e., the prevalence a perfect predictor with no error rate would have. Best score function for each dataset is presented in bold. Median across $100$ runs.}
\end{table}

On CIFAR10, where there was only one minority class, the prevalence remained low, with the lowest being $1.22\%$ achieved by the Mean Distance score function using cosine distance both in feature space and in logit space. On CIFAR100, despite that the minority classes only constituted about $10.5/10,000$ of examples each, they were each, on average, prevalent in $7\%-28\%$ of prediction regions, depending on the score function. Here, the lowest was again achieved by Mean Distance computed in feature space with cosine distance.

\FloatBarrier

\section{Discussion}
Label Distance, one of the simplest possible score functions, performed very well across a variety of datasets and models when using cosine distance as the distance metric. Margin Distance performed exceptionally well in logit space but consistently poorly on the probability simplex, and typically it performed better with cosine distance with the exceptions of EfficientNet on ImageNet and CIFAR10, as well as ViT on ImageNet. Mean Distance performed exceptionally well, and was the best choice for all models and all datasets apart from specifically the ResNet models on ImageNet. All versions of APS, especially when non-regularized, consistently performed better without the randomization term, letting $U$ instead be a small constant serving as a tie-breaker. Under this circumstance, SAPS typically performed better than RAPS which typically performed better than APS, though a notable exception is that APS performed the best of the three on CIFAR10, and this was consistent across the models. Overall, the APS algorithms had prediction set sizes slightly larger than the best distance-based score functions, however, we do note that only one set of parameters was used for these algorithms across all benchmarks and that they may not be at their respective theoretical optimum. The Gradient Distance and Fast Gradient Distance scores consistently performed well, with Fast Gradient Distance being the best-performing in the main analysis on ImageNet\footnote{In version v1 of this pre-print on arXiv, a bug caused the score of the Fast Gradient score function to be much higher in all evaluations.}. Furthermore, with the Fast Gradient Distance score consistently performing similarly to or better than Gradient Distance, there is not much reason to use the latter, significantly more computationally heavy, iterative approach.

A clear result from the benchmarking is that distance-based score functions measured with cosine distance perform very well in high-dimensional conformal domains. It is unsurprising that Euclidean distance is not well-suited for high-dimensional spaces, but the fact that cosine distance performs well suggests that these models are sorting classes into approximately orthogonal clusters. A notable exception is that Euclidean distance still performs similar to or better than cosine distance for Margin Distance in logit space for several models, even in the high-dimensional logit space of ImageNet. Finally, we note that Mean Distance even on the probability simplex frequently outperforms Label Distance, i.e., it is more fruitful to measure distance from the mean prediction of the model than to measure distance from the perfect prediction. In fact, using Table \ref{totaltable} we see that for Euclidean distance this was \textit{always} true.

ResNet50 and ResNet18 had nearly identical relative ordering of $I_1$-values across all cohorts. EfficientNet-B0 and ViT-B16 had highly similar relative orderings to each other, but different from those of the ResNet architectures. This is despite the fact that each model reached similar levels of convergence during training (appendix \ref{sec:appendixC}). This suggests that architecture may matter more than model performance in choosing score function. It is not obvious that model architecture should matter when the only conformal domains utilized are those in the final feature layer, logit vector layer, or output layer of the model, but subtle differences such as ResNet's usage of ReLU activation, which is identically zero for negative inputs, versus EfficientNet's usage of Swish activation and ViT's usage of GeLU activation which are not, may perhaps have influenced properties such as sparsity even in these final layers.

The results from the imbalanced CIFAR10 and CIFAR100 datasets suggest that substantial class imbalance proves a large challenge in conformal prediction. This challenge is largely inherited from the underlying predictive model with regard to accurately predicting rare classes. Furthermore, class imbalance reduces the precision of the conformal predictor due to a sparsity in calibration examples belonging to the rare class. Class-conditional conformal prediction is intended to provide a solution to precisely this challenge, but a conformal predictor may, as we see, ``solve" this by simply including the difficult classes in a very large number of prediction sets in order to achieve coverage for these classes. While this indeed damages the credibility of the prediction sets that include the rare classes, these are still the most ``honest" predictions, reflecting the fact that the underlying model has so much uncertainty with regard to the rare classes that it cannot state with high probability that a data point does not belong to this class. There are some recent works that address this. For instance, work by Ding et. al. involves compromising with class-conditional CP and obtaining coverage for clusters of classes \parencite{Ding2023} or maximizing a proposed ``macro-coverage" \parencite{macro_coverage_1, macro_coverage_2}. Another proposal involves using the rank of the class in model prediction to determine whether an example should be included in the class-conditional CP inference at all, or whether it should instead be discarded immediately due to poor rank \parencite{label_rank}. This method still achieves coverage by adapting the error rate on the subset that remains after filtering by rank.

This study has multiple strengths and limitations. A strength of this study is that it is the first study to use a metric designed to assess score functions themselves, and to use it to evaluate and compare score functions directly. Another strength is the introduction of a simple language surrounding score functions. A third strength is the clarification on the permissibility of utilizing any distance metric, with clear and novel examples. A limitation of this study is the lack of mechanistic investigation into the sample distributions in the conformal domains, which may reveal further insights into the circumstances under which score functions may perform particularly well or poorly. Another limitation is the lack of investigation into cases where the exchangeability assumption fails to hold, which is often the case in real-world applications. For instance, since the exchangeability is assumed between calibration and test data as opposed to the input images (or other data modality) themselves, the probability simplex may be a more robust conformal domain than feature space with regard to preserving exchangeability. We leave problems of investigating score functions in situations where exchangeability only holds approximately open for future research.

In this study, we have discussed and produced a benchmark for score functions in conformal prediction, in a machine learning classification setting. We have introduced the concept of the conformal domain and demonstrated how it can be the space of output probabilities, output logit vectors (prior to softmax), or model feature vectors. We have distinguished between distance-based and rank-based score functions, and shown how distance-based score functions can utilize non-euclidean distance metrics such as cosine distance. We have introduced an evaluation metric for score functions which is independent of an arbitrary choice of error rate. Further, the study has addressed a knowledge gap with regard to how to choose score functions when implementing conformal prediction in this setting. We have observed that no one score function can be ruled the best in the general case, but rather, different score functions perform well under different circumstances. Despite this, we contend that this benchmark provides a useful starting point for exploring which score functions to use, utilizing Table \ref{table1} and Table 3.

\section*{Code availability}
All functions used to generate the results in this paper are openly available on GitHub: \url{https://github.com/SolErikaB/conformal\_overview}. Generative AI coding assistants (Github copilot and Claude Sonnet) were utilized for large parts of the implementation.

\section*{Acknowledgments}
This project was made possible by computational resources funded by SciLifeLab \& Wallenberg Data Driven Life Science Program via a grant (KAW 2024.0159) to assistant prof. Kimmo Kartasalo, as well as the supercomputing resource Berzelius provided by the National Supercomputer Centre at Linköping University and the Knut and Alice Wallenberg Foundation. Thank you to Kimmo Kartasalo and to Magnus Boman for proof-reading the paper.

\printbibliography

\newpage
\begin{appendices}
\section{Complete table of $I_1$}
\label{sec:appendixA}
\addcontentsline{toc}{section}{Complete version of table}

\setcounter{figure}{0}
\renewcommand{\thefigure}{S\arabic{figure}}

\setcounter{table}{0}
\renewcommand{\thetable}{S\arabic{table}}

\setlength{\LTleft}{-3.7cm}  

\begin{longtable}{@{}lllllll@{}}
\\
\toprule
\multicolumn{1}{c}{Model} & \multicolumn{1}{c}{Dataset} & \multicolumn{1}{c}{\makecell[c]{Model \\ Accuracy}} & \multicolumn{1}{c}{Score function} & \multicolumn{1}{c}{\makecell[c]{Conformal domain}} &   \multicolumn{1}{c}{\makecell[c]{Distance\\metric}} & \multicolumn{1}{c}{$I_1$} \\ 
\midrule
\endfirsthead

\multicolumn{6}{c}{\tablename\ \thetable\ -- \textit{Continued from previous page}} \\
\toprule
\multicolumn{1}{c}{Model} & \multicolumn{1}{c}{Dataset} & \multicolumn{1}{c}{\makecell[c]{Model \\ Accuracy}} & \multicolumn{1}{c}{Score function} & \multicolumn{1}{c}{\makecell[c]{Conformal\\domain}} &   \multicolumn{1}{c}{\makecell[c]{Distance metric}} & \multicolumn{1}{c}{$I_1$} \\ 
\midrule
\endhead

\midrule
\multicolumn{6}{r}{\textit{Continued on next page}} \\
\endfoot

\bottomrule
\endlastfoot
ResNet50 & ImageNet & 76.15\% & Label Distance  & Probability space & Euclidean distance & 137.7\\
& &  & Label Distance  & Probability space & Cosine distance & 8.456 \\
& &  & Margin Distance & Probability space & Euclidean distance & 166.2 \\
& &  & Margin Distance & Probability space & Cosine distance & 124.6 \\
& &  & Margin Distance & Logit space       & Euclidean distance & 8.837 \\
&  &  & Margin Distance & Logit space & Cosine distance & 8.122 \\
&  &  & Mean Distance & Probability space       & Euclidean distance & 118.7 \\
&  &  & Mean Distance & Probability space       & Cosine distance & 8.527 \\
&  &  & Mean Distance & Logit space       & Euclidean distance & 40.66 \\
&  &  & Mean Distance & Logit space       & Cosine distance & 26.22 \\
&  &  & Mean Distance & Feature space       & Euclidean distance & 72.30 \\
&  &  & Mean Distance & Feature space       & Cosine distance & 18.58 \\
& &  & APS (U random)  & Probability space & N/A                & 20.85 \\
& &  & APS (U=0.001)   & Probability space & N/A                &  15.49\\
& &  & RAPS ($\lambda=0.2, k=1$, U random) & Probability space & N/A & 9.984\\
&  &  & RAPS ($\lambda=0.2, k=1$, U=0.001) & Probability space & N/A & 9.921\\
& &  & SAPS ($\lambda=0.2$, U random) & Probability space & N/A & 9.762 \\ 
& &  & SAPS ($\lambda=0.2$, U=0.001) & Probability space & N/A & 9.603 \\ 
 & &  & Fast Gradient & Features & Euclidean & 7.638 \\ 

\midrule
& CIFAR100 & 84.73\% & Label Distance  & Probability space & Euclidean distance & 13.89\\
& &  & Label Distance  & Probability space & Cosine distance & 5.568\\
& &  & Margin Distance & Probability space & Euclidean distance & 16.33 \\
&  &  & Margin Distance & Probability space & Cosine distance & 11.03\\
& &  & Margin Distance & Logit space       & Euclidean distance & 5.948 \\
& &  & Margin Distance & Logit space & Cosine distance & 5.352\\
&  &  & Mean Distance & Probability space       & Euclidean distance & 12.88 \\
&  &  & Mean Distance & Probability space       & Cosine distance & 3.049 \\
&  &  & Mean Distance & Logit space       & Euclidean distance & 4.451 \\
&  &  & Mean Distance & Logit space       & Cosine distance & 3.015 \\
&  &  & Mean Distance & Feature space       & Euclidean distance & 5.286 \\
&  &  & Mean Distance & Feature space & Cosine distance & 2.785 \\
& &  & APS (U random)  & Probability space & N/A                & 18.21 \\
& &  & APS (U=0.001)   & Probability space & N/A                &  6.709\\
& &  & RAPS ($\lambda=0.2, k=1$, U random) & Probability space & N/A & 5.025\\
&  &  & RAPS ($\lambda=0.2, k=1$, U=0.001) & Probability space & N/A & 4.922\\
& &  & SAPS ($\lambda=0.2$, U random) & Probability space & N/A & 5.015 \\ 
& &  & SAPS ($\lambda=0.2$, U=0.001) & Probability space & N/A & 4.754 \\ 
& &  & Gradient (lr=0.1, 100 steps) & Feature space & Euclidean Distance & 4.648\\ 
& &  & Fast Gradient & Feature space & Euclidean distance & 4.249 \\ 
\midrule
& CIFAR10 & 96.93\% & Label Distance  & Probability space & Euclidean distance & 1.362 \\
& &  & Label Distance  & Probability space & Cosine distance & 1.144\\
& &  & Margin Distance & Probability space & Euclidean distance & 1.398\\
&  &  & Margin Distance & Probability space & Cosine distance & 1.215\\
& &  & Margin Distance & Logit space       & Euclidean distance & 1.150 \\
& &  & Margin Distance & Logit space       & Cosine distance & 1.108 \\
&  &  & Mean Distance & Probability space       & Euclidean distance & 1.309 \\
&  &  & Mean Distance & Probability space       & Cosine distance & 1.116 \\
&  &  & Mean Distance & Logit space & Euclidean distance & 1.082\\
&  &  & Mean Distance & Logit space       & Cosine distance & 1.124 \\
&  &  & Mean Distance & Feature space       & Euclidean distance & 1.180 \\
&  &  & Mean Distance & Feature space       & Cosine distance & 1.278 \\
& &  & APS (U random)  & Probability space & N/A                & 1.576 \\
& &  & APS (U=0.001)   & Probability space & N/A                & 1.193 \\
& &  & RAPS ($\lambda=0.2, k=1$, U random) & Probability space & N/A & 1.386\\
&  &  & RAPS ($\lambda=0.2, k=1$, U=0.001) & Probability space & N/A & 1.350 \\
& &  & SAPS ($\lambda=0.2$, U random) & Probability space & N/A & 1.357 \\ 
& &  & SAPS ($\lambda=0.2$, U=0.001) & Probability space & N/A & 1.263 \\ 
& &  & Gradient (lr=0.1, 100 steps) & Feature space & Euclidean distance & 1.238 \\ 
& &  & Fast Gradient & Feature space & Euclidean distance & 1.124\\ 
 \bottomrule
ResNet18 & ImageNet & 69.76\% & Label Distance  & Probability space & Euclidean distance & 145.7\\
& &  & Label Distance  & Probability space & Cosine distance & 10.65 \\
& &  & Margin Distance & Probability space & Euclidean distance & 185.1 \\
& &  & Margin Distance & Probability space & Cosine distance & 143.6 \\
& &  & Margin Distance & Logit space       & Euclidean distance & 11.17 \\
&  &  & Margin Distance & Logit space & Cosine distance & 10.37 \\
&  &  & Mean Distance & Probability space       & Euclidean distance & 141.8 \\
&  &  & Mean Distance & Probability space       & Cosine distance & 10.92 \\
&  &  & Mean Distance & Logit space       & Euclidean distance & 51.94 \\
&  &  & Mean Distance & Logit space       & Cosine distance & 38.48 \\
&  &  & Mean Distance & Feature space       & Euclidean distance & 93.18 \\
&  &  & Mean Distance & Feature space       & Cosine distance & 27.36 \\
& &  & APS (U random)  & Probability space & N/A                & 22.25\\
& &  & APS (U=0.001)   & Probability space & N/A                &  16.82\\
& &  & RAPS ($\lambda=0.2, k=1$, U random) & Probability space & N/A & 14.16\\
&  &  & RAPS ($\lambda=0.2, k=1$, U=0.001) & Probability space & N/A & 14.08\\
& &  & SAPS ($\lambda=0.2$, U random) & Probability space & N/A & 13.87 \\ 
& &  & SAPS ($\lambda=0.2$, U=0.001) & Probability space & N/A & 13.69 \\ 
& &  & Fast Gradient & Feature space & Euclidean distance & 9.704\\ 
\midrule
& CIFAR100 & 81.50\% & Label Distance  & Probability space & Euclidean distance & 12.17\\
& &  & Label Distance  & Probability space & Cosine distance & 5.074\\
& &  & Margin Distance & Probability space & Euclidean distance & 16.20 \\
&  &  & Margin Distance & Probability space & Cosine distance & 11.27\\
& &  & Margin Distance & Logit space       & Euclidean distance & 5.487 \\
& &  & Margin Distance & Logit space & Cosine distance & 5.043\\
&  &  & Mean Distance & Probability space       & Euclidean distance & 13.93 \\
&  &  & Mean Distance & Probability space       & Cosine distance & 3.119 \\
&  &  & Mean Distance & Logit space       & Euclidean distance & 12.40 \\
&  &  & Mean Distance & Logit space       & Cosine distance & 4.106 \\
&  &  & Mean Distance & Feature space       & Euclidean distance & 14.47 \\
&  &  & Mean Distance & Feature space & Cosine distance & 3.477 \\
& &  & APS (U random)  & Probability space & N/A                & 17.84 \\
& &  & APS (U=0.001)   & Probability space & N/A                &  6.498\\
& &  & RAPS ($\lambda=0.2, k=1$, U random) & Probability space & N/A & 6.193\\
&  &  & RAPS ($\lambda=0.2, k=1$, U=0.001) & Probability space & N/A & 6.013\\
& &  & SAPS ($\lambda=0.2$, U random) & Probability space & N/A & 6.186\\ 
& &  & SAPS ($\lambda=0.2$, U=0.001) & Probability space & N/A & 5.811\\ 
 & &  & Gradient (lr=0.1, 100 steps) & Features & Euclidean & 4.201\\ 
& &  & Fast Gradient & Feature space & Euclidean distance & 4.219\\ 
\midrule
& CIFAR10 & 96.00\% & Label Distance  & Probability space & Euclidean distance & 1.330\\
& &  & Label Distance  & Probability space & Cosine distance & 1.177\\
& &  & Margin Distance & Probability space & Euclidean distance & 1.398\\
&  &  & Margin Distance & Probability space & Cosine distance & 1.280\\
& &  & Margin Distance & Logit space       & Euclidean distance & 1.188\\
& &  & Margin Distance & Logit space       & Cosine distance & 1.167\\
&  &  & Mean Distance & Probability space       & Euclidean distance & 1.285\\
&  &  & Mean Distance & Probability space       & Cosine distance & 1.313\\
&  &  & Mean Distance & Logit space & Euclidean distance & 1.123\\
&  &  & Mean Distance & Logit space       & Cosine distance & 1.162\\
&  &  & Mean Distance & Feature space       & Euclidean distance & 1.500\\
&  &  & Mean Distance & Feature space       & Cosine distance & 1.241\\
& &  & APS (U random)  & Probability space & N/A                & 2.082\\
& &  & APS (U=0.001)   & Probability space & N/A                & 1.187\\
& &  & RAPS ($\lambda=0.2, k=1$, U random) & Probability space & N/A & 1.368\\
&  &  & RAPS ($\lambda=0.2, k=1$, U=0.001) & Probability space & N/A & 1.320\\
& &  & SAPS ($\lambda=0.2$, U random) & Probability space & N/A & 1.405\\ 
& &  & SAPS ($\lambda=0.2$, U=0.001) & Probability space & N/A & 1.233\\ 
& &  & Gradient (lr=0.1, 100 steps) & Feature space & Euclidean distance & 1.156\\ 
& &  & Fast Gradient & Feature space & Euclidean distance & 1.144\\ 
 \bottomrule
EfficientNet-B0 & ImageNet & 77.67\% & Label Distance  & Probability space & Euclidean distance & 123.4\\
& &  & Label Distance  & Probability space & Cosine distance & 10.65 \\
& &  & Margin Distance & Probability space & Euclidean distance & 159.0 \\
& &  & Margin Distance & Probability space & Cosine distance & 109.1 \\
& &  & Margin Distance & Logit space       & Euclidean distance & 11.30 \\
&  &  & Margin Distance & Logit space & Cosine distance & 13.79 \\
&  &  & Mean Distance & Probability space       & Euclidean distance & 111.6 \\
&  &  & Mean Distance & Probability space       & Cosine distance & 7.358 \\
&  &  & Mean Distance & Logit space       & Euclidean distance & 557.3 \\
&  &  & Mean Distance & Logit space       & Cosine distance & 10.42 \\
&  &  & Mean Distance & Feature space       & Euclidean distance & 645.6 \\
&  &  & Mean Distance & Feature space       & Cosine distance & 20.26 \\
& &  & APS (U random)  & Probability space & N/A                & 49.00 \\
& &  & APS (U=0.001)   & Probability space & N/A                &  26.18\\
& &  & RAPS ($\lambda=0.2, k=1$, U random) & Probability space & N/A & 12.01\\
&  &  & RAPS ($\lambda=0.2, k=1$, U=0.001) & Probability space & N/A & 11.92\\
& &  & SAPS ($\lambda=0.2$, U random) & Probability space & N/A & 11.80 \\ 
& &  & SAPS ($\lambda=0.2$, U=0.001) & Probability space & N/A & 11.61 \\ 
& &  & Fast Gradient & Feature space & Euclidean distance & 9.677\\ 
\midrule
& CIFAR100 & 85.22\% & Label Distance  & Probability space & Euclidean distance & 13.24\\
& &  & Label Distance  & Probability space & Cosine distance & 3.792\\
& &  & Margin Distance & Probability space & Euclidean distance & 15.37 \\
&  &  & Margin Distance & Probability space & Cosine distance & 9.421\\
& &  & Margin Distance & Logit space       & Euclidean distance & 3.975 \\
& &  & Margin Distance & Logit space & Cosine distance & 3.930\\
&  &  & Mean Distance & Probability space       & Euclidean distance & 11.12\\
&  &  & Mean Distance & Probability space       & Cosine distance & 2.422 \\
&  &  & Mean Distance & Logit space       & Euclidean distance & 17.56\\
&  &  & Mean Distance & Logit space       & Cosine distance & 3.154 \\
&  &  & Mean Distance & Feature space       & Euclidean distance & 35.44\\
&  &  & Mean Distance & Feature space & Cosine distance & 2.307 \\
& &  & APS (U random)  & Probability space & N/A                & 12.45\\
& &  & APS (U=0.001)   & Probability space & N/A                &  5.803\\
& &  & RAPS ($\lambda=0.2, k=1$, U random) & Probability space & N/A & 4.316\\
&  &  & RAPS ($\lambda=0.2, k=1$, U=0.001) & Probability space & N/A & 4.243\\
& &  & SAPS ($\lambda=0.2$, U random) & Probability space & N/A & 4.240\\ 
& &  & SAPS ($\lambda=0.2$, U=0.001) & Probability space & N/A & 4.046\\ 
& &  & Gradient (lr=0.1, 100 steps) & Feature space & Euclidean distance & 3.261\\ 
& &  & Fast Gradient & Feature space & Euclidean distance & 3.271\\ 
\midrule
& CIFAR10 & 97.09\% & Label Distance  & Probability space & Euclidean distance & 1.306\\
& &  & Label Distance  & Probability space & Cosine distance & 1.131\\
& &  & Margin Distance & Probability space & Euclidean distance & 1.345\\
&  &  & Margin Distance & Probability space & Cosine distance & 1.193\\
& &  & Margin Distance & Logit space       & Euclidean distance & 1.136\\
& &  & Margin Distance & Logit space       & Cosine distance & 1.110\\
&  &  & Mean Distance & Probability space       & Euclidean distance & 1.239\\
&  &  & Mean Distance & Probability space       & Cosine distance & 1.073\\
&  &  & Mean Distance & Logit space & Euclidean distance & 1.139\\
&  &  & Mean Distance & Logit space       & Cosine distance & 1.194\\
&  &  & Mean Distance & Feature space       & Euclidean distance & 8.379\\
&  &  & Mean Distance & Feature space       & Cosine distance & 1.130\\
& &  & APS (U random)  & Probability space & N/A                & 1.754\\
& &  & APS (U=0.001)   & Probability space & N/A                & 1.239\\
& &  & RAPS ($\lambda=0.2, k=1$, U random) & Probability space & N/A & 1.402\\
&  &  & RAPS ($\lambda=0.2, k=1$, U=0.001) & Probability space & N/A & 1.368\\
& &  & SAPS ($\lambda=0.2$, U random) & Probability space & N/A & 1.388\\ 
& &  & SAPS ($\lambda=0.2$, U=0.001) & Probability space & N/A & 1.278\\ 
& &  & Gradient (lr=0.1, 100 steps) & Feature space & Euclidean distance & 1.131\\ 
& &  & Fast Gradient & Feature space & Euclidean distance & 1.112\\ 
 \bottomrule
ViT-B16 & ImageNet & 81.07\% & Label Distance  & Probability space & Euclidean distance & 134.5\\
& &  & Label Distance  & Probability space & Cosine distance & 8.930 \\
& &  & Margin Distance & Probability space & Euclidean distance & 158.7 \\
& &  & Margin Distance & Probability space & Cosine distance & 101.7 \\
& &  & Margin Distance & Logit space       & Euclidean distance & 9.537 \\
&  &  & Margin Distance & Logit space & Cosine distance & 22.34 \\
&  &  & Mean Distance & Probability space       & Euclidean distance & 110.7 \\
&  &  & Mean Distance & Probability space       & Cosine distance & 6.881 \\
&  &  & Mean Distance & Logit space       & Euclidean distance & 670.1 \\
&  &  & Mean Distance & Logit space       & Cosine distance & 8.806\\
&  &  & Mean Distance & Feature space       & Euclidean distance & 485.9 \\
&  &  & Mean Distance & Feature space       & Cosine distance & 15.05 \\
& &  & APS (U random)  & Probability space & N/A                & 40.87\\
& &  & APS (U=0.001)   & Probability space & N/A                &  25.85\\
& &  & RAPS ($\lambda=0.2, k=1$, U random) & Probability space & N/A & 11.44\\
&  &  & RAPS ($\lambda=0.2, k=1$, U=0.001) & Probability space & N/A & 11.39\\
& &  & SAPS ($\lambda=0.2$, U random) & Probability space & N/A & 11.23 \\ 
& &  & SAPS ($\lambda=0.2$, U=0.001) & Probability space & N/A & 11.11 \\ 
 & &  & Fast Gradient & Features & Euclidean & 8.392 \\ 
\midrule
& CIFAR100 & 86.74\% & Label Distance  & Probability space & Euclidean distance & 14.45\\
& &  & Label Distance  & Probability space & Cosine distance & 6.078\\
& &  & Margin Distance & Probability space & Euclidean distance & 15.59\\
&  &  & Margin Distance & Probability space & Cosine distance & 10.05\\
& &  & Margin Distance & Logit space       & Euclidean distance & 6.281 \\
& &  & Margin Distance & Logit space & Cosine distance & 5.889\\
&  &  & Mean Distance & Probability space       & Euclidean distance & 12.53\\
&  &  & Mean Distance & Probability space       & Cosine distance & 3.446 \\
&  &  & Mean Distance & Logit space       & Euclidean distance & 5.438\\
&  &  & Mean Distance & Logit space       & Cosine distance & 3.627 \\
&  &  & Mean Distance & Feature space       & Euclidean distance & 8.416 \\
&  &  & Mean Distance & Feature space & Cosine distance & 5.354 \\
& &  & APS (U random)  & Probability space & N/A & 12.24\\
& &  & APS (U=0.001)   & Probability space & N/A &  7.257\\
& &  & RAPS ($\lambda=0.2, k=1$, U random) & Probability space & N/A & 6.187\\
&  &  & RAPS ($\lambda=0.2, k=1$, U=0.001) & Probability space & N/A & 6.155\\
& &  & SAPS ($\lambda=0.2$, U random) & Probability space & N/A & 6.106\\ 
& &  & SAPS ($\lambda=0.2$, U=0.001) & Probability space & N/A & 5.997\\ 
& &  & Gradient (lr=0.1, 100 steps) & Feature space & Euclidean distance & 5.874\\ 
& &  & Fast Gradient & Feature space & Euclidean distance & 5.140\\ 
\midrule
& CIFAR10 & 97.66\% & Label Distance  & Probability space & Euclidean distance & 1.600\\
& &  & Label Distance  & Probability space & Cosine distance & 1.198\\
& &  & Margin Distance & Probability space & Euclidean distance & 1.628\\
&  &  & Margin Distance & Probability space & Cosine distance & 1.249\\
& &  & Margin Distance & Logit space       & Euclidean distance & 1.202\\
& &  & Margin Distance & Logit space       & Cosine distance & 1.086\\
&  &  & Mean Distance & Probability space       & Euclidean distance & 1.406\\
&  &  & Mean Distance & Probability space       & Cosine distance & 1.108\\
&  &  & Mean Distance & Logit space & Euclidean distance & 1.137\\
&  &  & Mean Distance & Logit space       & Cosine distance & 1.078\\
&  &  & Mean Distance & Feature space       & Euclidean distance & 3.379\\
&  &  & Mean Distance & Feature space       & Cosine distance & 1.426\\
& &  & APS (U random)  & Probability space & N/A                & 1.467\\
& &  & APS (U=0.001)   & Probability space & N/A                & 1.276\\
& &  & RAPS ($\lambda=0.2, k=1$, U random) & Probability space & N/A & 1.406\\
&  &  & RAPS ($\lambda=0.2, k=1$, U=0.001) & Probability space & N/A & 1.360\\
& &  & SAPS ($\lambda=0.2$, U random) & Probability space & N/A & 1.358\\ 
& &  & SAPS ($\lambda=0.2$, U=0.001) & Probability space & N/A & 1.276\\ 
& &  & Gradient (lr=0.1, 100 steps) & Feature space & Euclidean distance & 1.306\\ 
& &  & Fast Gradient & Feature space & Euclidean distance & 1.174\\ 
 \bottomrule
 
\caption{Values of $I_1$, defined in (\ref{integral2}), for different score functions, conformal domains, and distance metrics. Lower is better.}
\label{totaltable}
\end{longtable}

\FloatBarrier

\section{Equivalence of APS and RAPS algorithms with our standard conformal prediction procedure}
\label{sec:appendixB}
The APS algorithm as originally described by \textcite{romanoClassificationValidAdaptive2020a}, section 2.2, is as follows (not verbatim, language and symbols are adapted to match our previous descriptions):

\begin{enumerate}
    \item Given a predictive classification model and a calibration data set $\mathbb{C}$ not seen during training, denote by $X_i$ the i-th data point and $y_i$ its class (i.e., its label). Denote by $\hat{\pi}_i$ the ordered list of probabilities for the i-th example.
    \item Sample a uniform random variable $U_i \sim \text{Uniform}(0,1)$ for each example.
    \item Define the (non-conformity) score function
    \begin{equation} \label{aps_score_supp}
       s(y_i, U_i, \hat{\pi}_i) = \min\{\tau: y_i \in \mathcal{S}_i(U_i, \hat{\pi}_i, \tau)\},
   \end{equation}
   \begin{equation}
   \label{eq:APS_S}
    \quad \mathcal{S}_i(U_i, \hat{\pi}_i, \tau)  = \begin{cases} \{y_i^{(j)} : j < \min{\{j' \ | \ \sum_{k=1}^{j'} \hat{\pi}^{(k)}_i \geq \tau \} \} } & \text{if}\ U_i \leq V_i(\hat{\pi}_i, \tau) \\ \{y_i^{(j)} : j \leq \min{\{j' \ | \ \sum_{k=1}^{j'} \hat{\pi}^{(k)}_i \geq \tau \} \} } & \text{if}\ U_i>V_i(\hat{\pi}_i, \tau) \end{cases},
    \end{equation}                                                                                  
    \begin{equation}
    V_i(\hat{\pi}_i, \tau) = \frac{1}{\hat{\pi}_i^{(c)}}(\sum_{\hat{\pi}^{(j)} \geq \hat{\pi}^{(c)}}{\hat{\pi}^{(j)} - \tau}).
    \end{equation}
    Here, $y_i$ is the label of the i-th example, $y_i^{(j)}$ is the class that has index $j$ from the ordered list of predicted probabilities belonging to example $i$, and $\hat{\pi}_i^{(c)}$ is the predicted probability of the true class for example $i$.
    \item Calibrate the conformal predictor by calculating $s_i$ for all the examples in $\mathbb{C}$.
    \item Choose an allowed error rate (sometimes called significance) $\alpha$.
    \item Construct the prediction set for a new test example as $\mathcal{S}_{i+1}(U_{i+1}, \hat{\pi}_{i+1}, \tau_c)$ with $\tau_c$ the $\frac{\lceil (|\mathbb{C}|+1)(1-\alpha) \rceil}{|\mathbb{C}|}$-th quantile of $s_i$ on $\mathbb{C}$.
\end{enumerate}

A naïve implementation of the above algorithm in pseudo-code could look like

\lstset{style=mystyle}

\begin{lstlisting}

def calibration
    vars label, sorted_indices, sorted_scores
    
    c = cumsum(sorted_scores)
    
    while calibrating:
        L = index(sorted_indices==label)
        U = uniform(0,1)
        for possible_tau in range(0,1,0.0001):
            V = (c[L]-possible_tau)/sorted_scores[L]
            if U<=V: new_L = L-1
            else if U>V: new_L = L
            prediction_set = sorted_indices[0,1,2,...,new_L]
            if label in prediction_set:
                return possible_tau

for each example:
    tau_list add calibration(label, sorted_indices, sorted_scores)

threshold = quantile(tau_list, error_rate_with_finite_sample_adjustment)

def testing
    vars sorted_indices, sorted_scores, threshold

    c = cumsum(sorted_scores)
    U = uniform(0,1)
    L = index(c > threshold)
    V = (c[L]-threshold)/sorted_scores[L]

    if U<=V: return sorted_indices[1,2,...,L-1]
    else if U>V: return sorted_indices[1,2,...,L]
\end{lstlisting}

i.e. iterating over possible values of the threshold $\tau$ and stopping when the expected class becomes part of the prediction set. In practice, during calibration, the threshold that gets returned will always be the first tested threshold such that $U>V$. This is because when $U=V$ we have
\begin{equation}
    \mathcal{S} = \{y^{(j)} : j < \min{\{j' \ | \ \sum_{k=1}^{j'} \hat{\pi}^{(k)} \geq \tau \} \}},
\end{equation}
but
\begin{equation}
    \tau = \sum_{\hat{\pi}^{(j)} \geq \hat{\pi}^{(c)}}\hat{\pi}^{(j)}-U*\hat{\pi}^{(c)},
\end{equation}
so we have
\begin{equation}
    \mathcal{S} = \{y^{(j)} : j < \min{\{j' \ | \ \sum_{k=1}^{j'} \hat{\pi}^{(k)} \geq \sum_{\hat{\pi}^{(j)} \geq \hat{\pi}^{(c)}}{\hat{\pi}^{(j)}-U*\hat{\pi}^{(c)} \}}}.
\end{equation}
Further, we have that
\begin{equation}
    \sum_{k=1}^{c} \hat{\pi}^{(k)} > \sum_{\hat{\pi}^{(j)} \geq \hat{\pi}^{(c)}}{\hat{\pi}^{(j)}-U*\hat{\pi}^{(c)}},
\end{equation}
and hence the true class is never in $\mathcal{S}$ since it is not inclusive. However, as $U$ exceeds $V$, it becomes inclusive. Similarly, we have that
\begin{equation}
    \sum_{k=1}^{c-1} \hat{\pi}^{(k)} < \sum_{\hat{\pi}^{(j)} \geq \hat{\pi}^{(c)}}{\hat{\pi}^{(j)}-U*\hat{\pi}^{(c)}},
\end{equation}
whence the true class is always the next in line when $\mathcal{S}$ becomes inclusive, and hence it is always at this point that the minimum $\tau$ is found. This inspires us to invent a more clever alternative to the naïve algorithm, in which we replace step 3 with:
\begin{equation} \label{APS_score}
    s(y, U, \hat{\pi}_i) = \sum_{\hat{\pi}^{(j)} \geq \hat{\pi}^{(c)}}\hat{\pi}^{(j)}-U*\hat{\pi}^{(c)},
\end{equation}
or equivalently, redrawing the uniform variable (and changing $U \rightarrow 1-U$ in \ref{eq:APS_S}):
\begin{equation}
    s(y, U, \hat{\pi}_i) = \sum_{\hat{\pi}^{(j)} > \hat{\pi}^{(c)}}\hat{\pi}^{(j)}+U*\hat{\pi}^{(c)}
\end{equation}
This is the form presented by \textcite{angelopoulosUncertaintySetsImage2020a}. However, in that paper, during inference the prediction sets are still formed without including the $U*\hat{\pi}^{c}$ term which is compensated for by random discarding with $V$. Here, we compute $s$ similarly during calibration and inference, removing all need for random discarding.\\

The APS algorithm does have an advantage over the algorithm $\ref{subsec:CPalg}$. Because the non-conformity score itself is computed by creating prediction sets and asking which threshold would lead to the true class being included, the final prediction set for new test examples is given by simply including all classes in descending order based on their predicted probabilities until the cumulative sum exceeds the calibrated threshold, without requiring computing separate cumulative sums for each possible class individually. The prediction sets are trivially the same as one gets if one applies the algorithm described in section $\ref{subsec:CPalg}$, but the time-complexity of the algorithm no longer scales with the number of classes in the underlying dataset.

\section{Training specifications}
\label{sec:appendixC}

Here we describe the fine-tuning of the resnet18, resnet50, efficientnet\_b0, and ViT-B16, (all pre-trained on ImageNet) on CIFAR10 and CIFAR100.

The implementation was done using pytorch. The torchvision.transforms library was used to apply random cropping (RandomCrop(32, padding=4)), random flipping (RandomHorizontalFlip()), and normalization (Normalize()), on the training data. The validation data used to check that the models were adequately trained was not augmented apart from normalization.

Images were upsampled from 32x32 to 224x224 to work with the ImageNet pre-trained weights. The final layer of each model was replaced with one whose output size corresponded to the number of classes of the corresponding dataset.

Training utilized a cross entropy loss function (torch.nn.CrossEntropyLoss(label\_smoothing=0.1)) and an SGD optimizer (torch.optim.SGD(lr=0.01, momentum=0.9, weight\_decay=5e-4)). Learning rate used a cosine annealing schedule (optim.lr\_scheduler.CosineAnnealingLR(optimizer, T\_max=50)). Training was done for 50 epochs regardless of validation accuracy. In Supplementary Figure 1, we show that despite that 50 epochs was likely more than necessary, there were no issues with overfitting.

When simulating class imbalance, the above settings were again used, but with 10\% of the classes reduced to 10\% of the samples. Both the classes as well as which data points in those classes were chosen randomly. 

\begin{figure}
    \centering
    \includegraphics[width=\linewidth]{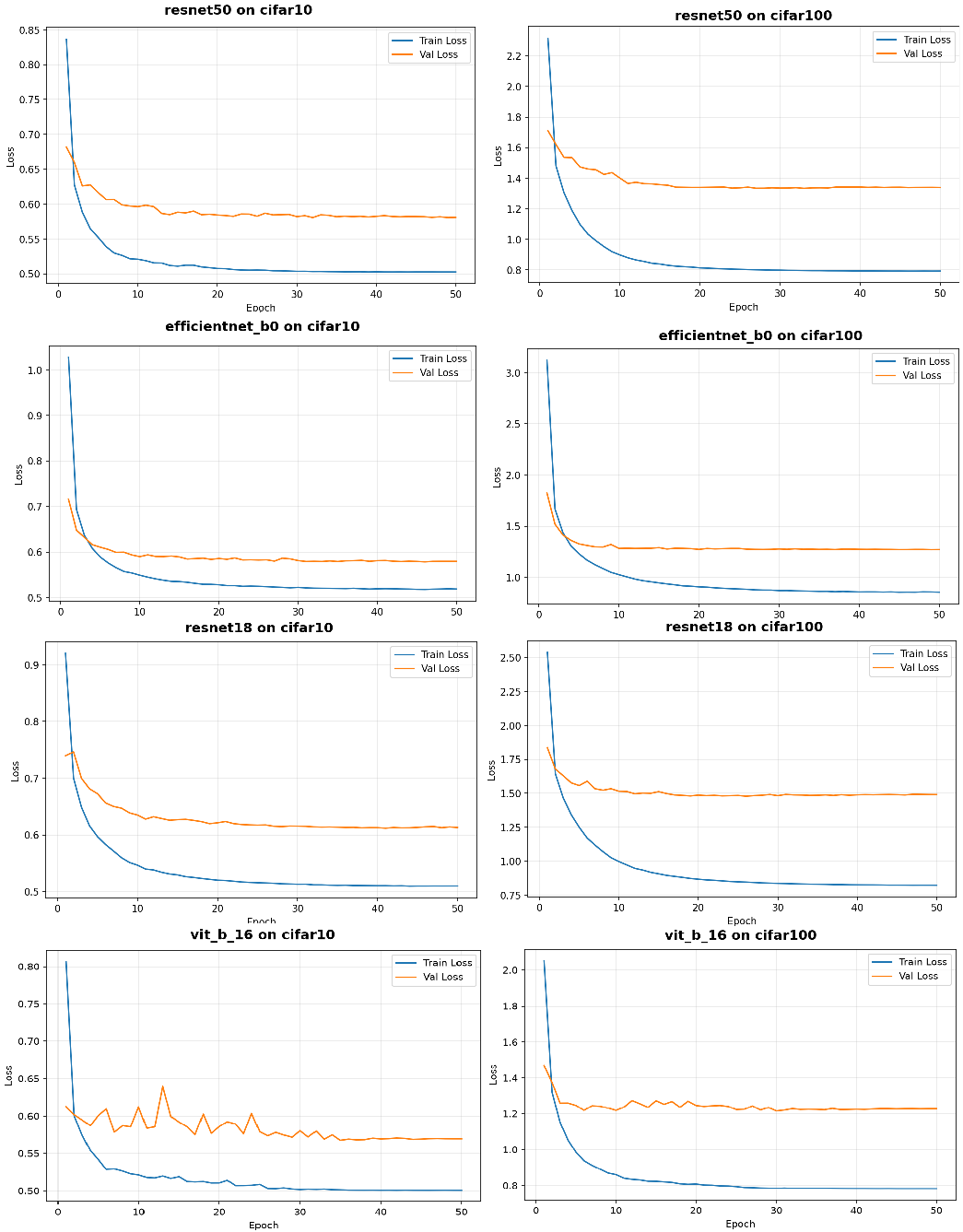}
    \caption{Training and validation loss during fine-tuning of the ResNet50, ResNet18, EfficientNet-B0 and ViT-B16 on the CIFAR10 and CIFAR100 datasets.}
    \label{fig:training_curves}
\end{figure}

\end{appendices}

\end{document}